\documentclass{article} 
\usepackage{iclr2016_conference,times}
\usepackage{hyperref}
\usepackage{url}
\usepackage{times}
\usepackage{epsfig}
\usepackage{graphicx}
\usepackage{amsmath}
\usepackage{amssymb}
\usepackage{subcaption}
\usepackage{color}
\usepackage{float}
\usepackage{footnote}
\usepackage{enumitem}
\makesavenoteenv{tabular}
\makesavenoteenv{table}

\title{Local Handcrafted Features are Convolutional Neural Networks}

\author{Zhenzhong Lan, Shoou-I Yu, Ming Lin,  Bhiksha Raj, Alexander G. Hauptmann  \\
School of Computer Science,
Carnegie Mellon University,
Pittsburgh, PA 15213, USA \\
\texttt{\{lanzhzh, iyu, minglin, bhiksha, alex\}@cs.cmu.edu} \\
}

\begin{document}

\maketitle

\begin{abstract}

Image and video classification research has made great progress through the development of handcrafted local features and learning based features. These two architectures were proposed roughly at the same time and have flourished at overlapping stages of history. However, they are typically viewed as distinct approaches. In this paper, we emphasize their structural similarities and show how such a unified view helps us in designing features that balance efficiency and effectiveness. As an example, we study the problem of designing efficient video feature learning algorithms for action recognition. 

We approach this problem by first showing that local handcrafted features and Convolutional Neural Networks (CNNs) share the same convolution-pooling network structure. We then propose a two-stream Convolutional ISA (ConvISA) that adopts the convolution-pooling  structure of the state-of-the-art handcrafted video feature with greater modeling capacities and a cost-effective training algorithm. Through custom designed network structures for pixels and optical flow, our method also reflects distinctive characteristics of these two data sources. 

Our experimental results on standard action recognition benchmarks show that by focusing on the structure of CNNs, rather than end-to-end training methods,  we are able to design an efficient and powerful video feature learning algorithm.

\end{abstract}

\section{Introduction}

Despite a long history of prior work, image and video classification remains challenging. Recent progress on this problem mainly relies on the improvements of features, which can be broadly categorized into two classes: 1) local handcrafted approaches  (\cite{schmid1997local,lowe2004distinctive, wang2011action}) that include handcrafted features and their corresponding encoding methods (\cite{lazebnik2006beyond, perronnin2010improving}), and 2) learning based approaches that are mainly represented by CNNs for image recognition (\cite{lecun1998gradient, krizhevsky2012imagenet, simonyan2014two}). These two groups of methods have strong connections in both functionality and history but are typically treated as unrelated methodologies. 

In a ground-breaking work, \cite{schmid1997local} proposed to use local features for general image retrieval. Since then, handcrafted features have become very popular and numerous improvements have sprung up (\cite{lowe2004distinctive, fei2005bayesian, perronnin2010improving, wang2011action}. They have been the best practices for image and video classification until recently. In particular, SIFT based features with Fisher Vector encoding were still the best ingredients for PASCAL VOC challenges 2012 (\cite{everingham2014pascal}). The reason behind this popularity is that handcrafted features do not rely on any labeled data and have very efficient training algorithms. The main problem of these methods, however, is that their modeling capacities are limited by the fixed transformations (filters) that stay the same for different sources of data. For example, HOG (\cite{dalal2005histograms}) and HOF (\cite{laptev2005space}) use the same set of gradient and binning filters despite different inputs. 

Roughly at the same time of \cite{schmid1997local}'s work, \cite{lecun1998gradient} published the first modern CNN architecture (LeNet-5) for handwritten character recognition. The three key aspects of this CNN framework are the convolution-pooling architecture of the model, the operational functions that implement this architecture, and the objective function
that is optimized to learn the parameters of these functions.  All three contributes to the {\em effectiveness} of the features obtained from the framework.  Of the three, the latter two, namely the operational functions and the optimization criterion have the greater influence on the data requirements to learn the model and the computational expense of the learning. In fact, these were the major bottleneck for extending LetNet-5 from dealing with simple image processing tasks like digital recognition to processing more complex ones such as object recognition from natural images. 

However, the first aspect, namely the convolution-pooling architecture too has a significant influence on the effectiveness of
the features. Indeed, as we show in Section \ref{sec:architecture}, state-of-the-art handcrafted features for image and video classification do implicitly employ this architecture.

We therefore retain only the first aspect, namely the conovlutional-pooling architecture, but replace the operational
functions and optimization objective with simpler, easier to compute variants. As a case in point, we design a cost-effective video feature learning method based on the conovlutional-pooling architecture. Video feature learning is chosen because it often involves larger data size and more difficult labeling than image feature learning. For these kind of tasks, CNNs face the same problems as they had for object recognition a decade ago. 

For restricting the computational complexity and staying label-free, we adopt the architecture of Improved Dense Trajectory (IDT) (\cite{wang2013action}), a state-of-the-art local video feature. To avoid the modeling capacity limitation of handcrafted transformation, we replace them with a Convolutional ISA model. Our proposed architecture, termed two-stream ConvISA, has the following attractive properties:
\begin{itemize}[noitemsep,nolistsep]
\item It has the convolution-pooling architecture that shares among handcrafted and CNN approaches. 
\item Compared to handcrafted approaches, it has much more powerful modeling capabilities and can be adapted to different data sources. 
\item Compared to end-to-end CNN approaches, it avoids the costly label collection and model training. 
\end{itemize}

We conduct experiments on benchmark action datasets of HMDB51 and UCF101, as in \cite{simonyan2014two}. Table \ref{tab:comparison} compares the model training time and accuracy of our method to the two-stream CNN (\cite{simonyan2014two}), a state-of-the-art CNN approach. Two-stream CNNs are trained in a multi-task learning manner by utilizing both UCF101 and HMDB51 datasets. Hence their results are not directly comparable to ours, but we put them here for reference purposes. In terms of training time, our approach is much more efficient than two-stream CNNs by several orders of magnitude. Two-stream CNNs needs about 1 day to train a model on 4 Titan GPUs while two-stream ConvISA only needs about 2 hours on 1 CPU. With regard to accuracy, our method outperforms two-stream CNNs on HMDB51 and has similar results on UCF101 despite the fact that it was trained on less data and does not need any labels to train the feature extraction module.   

\begin{table}[]
\centering
\begin{tabular}{|c|c|c| c | c|}
\hline
                    &Training Time & Need Label &  HMDB51 & UCF101 \\
                    \hline
Ours   & $\sim$2 hours / 1 CPU  &  No         &  \textbf{61.5\%} & \textbf{88.3\%}  \\
Two-stream CNNs     & $\sim$ 1 day / 4 GPUs & Yes   &  59.4\%   & 88.0\%  \\
  \hline
\end{tabular}
\caption{Comparison of our approach with two-stream CNNs.}
\label{tab:comparison}
\end{table}

In the remainder of this paper, we first provide more background information about video features with an emphasis on recent attempts at learning with deep neural networks. We then describe the relationship between handcrafted features and CNN-based features in detail, followed by the descriptions of our two-stream ConvISA algorithm. After that, we conduct experiments and show more detailed comparisons of our method to the state-of-the-art methods. Further discussions including potential improvements are provided at the end.

\section{Related Work}

Features and encoding are the major sources of breakthroughs in conventional video representations. Among them the trajectory based approaches (\cite{wang2013action,jiang2012trajectory}), especially the Dense Trajectory (DT) and IDT (\cite{wang2011action, wang2013action}), are the basis of current state-of-the-art handcrafted algorithms. These trajectory-based methods are designed to address the flaws of image-extended video features. Their superior performance validates the need for a unique representation of motion features.

There have been many works attempting to improve IDT due to its popularity.  \cite{peng2014bag} enhanced the performance of IDT by increasing codebook sizes and fusing multiple coding methods.  \cite{sapienza2014feature} explored ways to sub-sample and generate vocabularies for DT features.  \cite{hoai2014improving} achieved state-of-the-art performance on several action recognition datasets through applying three techniques including data augmentation, modeling score distribution over video subsequences, and capturing the relationship among action classes. \cite{fernandomodeling} modeled the evolution of appearance in the video and achieved state-of-the-art results on the Hollywood2 dataset. \cite{lan2014beyond} proposed to extract features from videos with multiple playback speeds to achieve speed invariances. However, none of them dealt with the fact that IDT relies on very simple handcrafted descriptors. In contrast, data-driven approaches have demonstrated their modeling power over image recognition (\cite{krizhevsky2012imagenet}) and are gradually replacing traditional handcrafted methods.

Motivated by the success of CNNs, researchers are working intensely towards developing CNN equivalents for learning video features. Several accomplishments have been reported from using CNNs for action recognition in videos (\cite{ wu2015modeling, varadarajan2015efficient}). \cite{karpathy2014large} trained deep CNNs through one million weakly labeled YouTube videos and reported moderate success while using it as a feature extractor.  \cite{simonyan2014two} demonstrated a result competitive to IDT (\cite{wang2013action}) through training deep CNNs using both sampled frames and stacked optical flows. \cite{tran2014c3d} developed C3D to extend 2D convolution to 3D convolution. \cite{ng2015beyond} proposed to to build a LSTM on top of CNNs to capture longer-term relationship among frames. \cite{wang2015action} use the outputs of two-stream CNNs as features to replace HOG and achieve state-of-the-art results on HMDB51 and UCF101 datasets. All of the above relied on a large amount of labels which are expensive to get and generally perform worse than handcrafted features among small datasets.

There has been limited number of works on unsupervised methods for learning video features. Among them the Independent Component Analysis (ICA) (\cite{hurri2003simple}) was the first approach to learn representations of videos in an unsupervised way.  \cite{le2011learning} addressed the issue using stacked ConvISA. \cite{srivastava2015unsupervised} applied unsupervised feature learning through long-short term memory. Since these methods rely purely on pixel data, they struggled to capture motion information and generally performed no better than state-of-the-art handcrafted methods. Our previous work (\cite{lan2015best}) tried to use stacked ConvISA to learn filters for optical flow data, but also could not outperform handcrafted methods due to the fact that stacked ConvISA does not have a good network structure for optical-flow feature learning. 

There are also several attempts at connecting the traditional feature encoding pipeline to the neural network frameworks. \cite{sydorov2014deep} studied the structure similarities between Fisher vectors and neural networks and proposed to jointly optimize Fisher vectors and the classifier.  \cite{richardbow}  converted the kMeans-based BoW model into an equivalent recurrent neural network and trained the BoW model and classifier together. Both above approaches focus on the end-to-end training of CNNs and again require labels and significantly increase the model training time. Instead, we emphasize the convoltional-pooling structure of CNNs rather than their training methods.

\section{The Convolution-Pooling Architecture}
\label{sec:architecture}
In this section we first define the convolution-pooling structure and then compare IDT with CNN-based video features. We highlight their structural similarities by showing that they are both features generated by deep convolution-pooling cascade with two key elements: convolution and pooling layers.

We define a convolution-pooling cascade as any single, iterative or
recursive implementation of the following sequence of operations:
\[
c(x) =  f(w \otimes x)
\]
\[
p(x) =  g(c(x))
\]
where  $w \otimes x$ is a three-dimensional convolution of a {\em
filter} $w$ with the $N\times M \times T$ video blocks $x$ and $f()$ is any non-linear
compressive operation.  $w \otimes x$, and as a consequence $c(x)$
also have size $N \times M \times T$.  $g()$ is a {\em pooling} function that
results in a shrinking of the argument and operates on any $N\times M \times T$
input to generate a $J \times K \times L $ output $p(x)$, where $J \leq N$, $K \leq
M$, and $ L \leq M$.

\subsection{Handcrafted Video Features}

\begin{figure}
\centering
\includegraphics[width=0.90\textwidth]{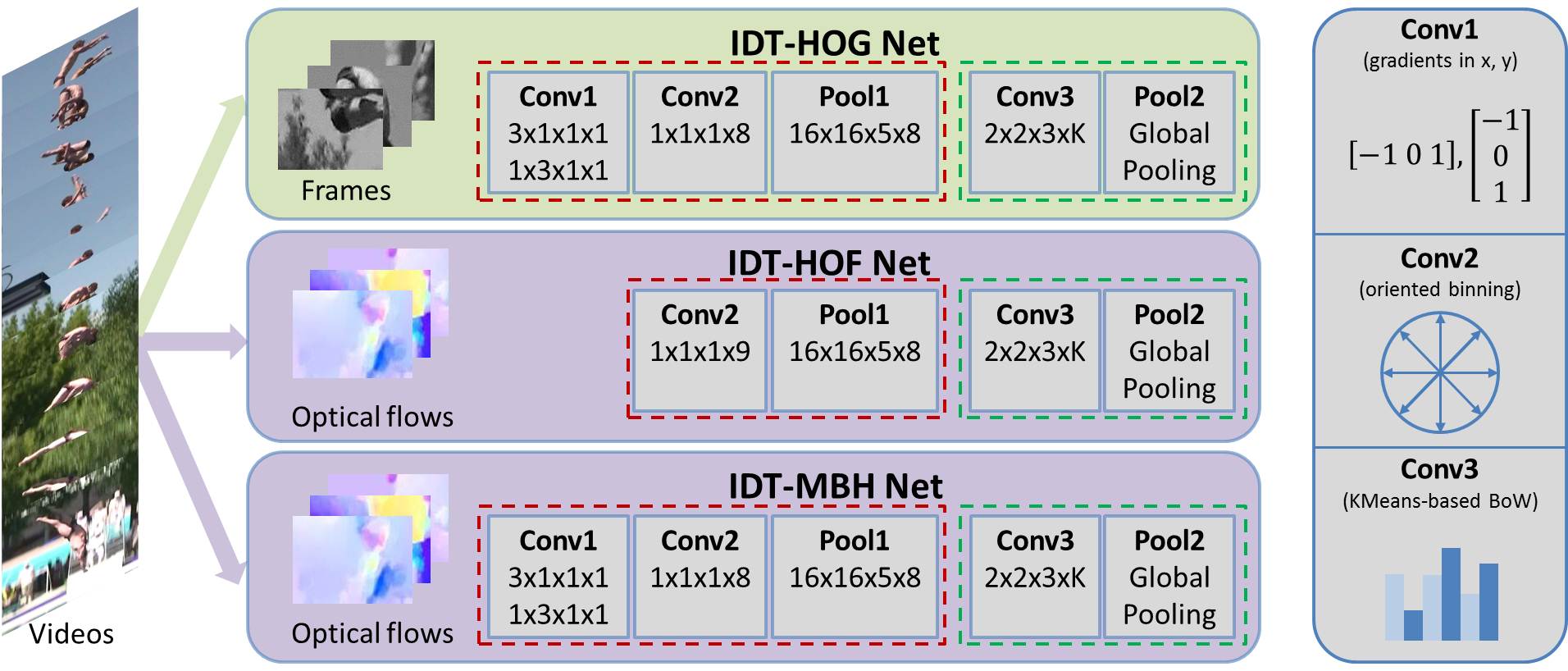}
\caption{ Schematic description of IDT as procedures of multiple convolution and pooling operations. Dashed \textcolor{red}{red} and \textcolor{green}{green} boxes represent the procedure of generating handcrafted descriptors and KMeans-based BoW encoding, respectively. In each operation, the first three numbers are the receptive field sizes in space and time (x, y, t) and the last number indicates the size of output channels. }
\label{fig:idt_net}
\end{figure}

A typical handcrafted video feature extraction procedure is often composed of two stages of convolution and pooling. The first stage purely relies on handcrafted filters and generates descriptors from local data. The second one often uses filters learned from unsupervised methods to encode the descriptors generated from the first stage and pool them together to get global features. For example, shown in Figure \ref{fig:idt_net} is a schematic description of three HOG-based IDT descriptors, each of which contains two stages of convolution and pooling (marked by dashed \textcolor{red}{red} and \textcolor{green}{green} boxes, respectively) including a total of  three convolution and two pooling operations. Among them, \textbf{Conv1} uses two gradient filters as $w$ and with: 
\[
f(x) = x.
\]
\textbf{Conv2} is an oriented soft binning, which can be approximated with $w$ being the unit directional vectors and $f$ being non-linear activation functions such as rectified linear unit  (\cite{krizhevsky2012imagenet}):
\[
f(x) = \max(x, 0).
\]
\textbf{Conv3} is KMeans-based BoW, which uses KMeans centroids as $w$ and a softmax function (\cite{richardbow}) as $f$: 
\[
f_k(x) = \frac{\exp(x_k)}{\sum_j \exp(x_j)},
\]
where $k$ is the $k$th centroid. \textbf{Pool1} is a local sum pooling: 
\[
g_{x,y,t}(x) = \sum_{j,l,m \in [1,d]}x_{xj,yl,tm },
\]
where $d$ is the pool size in space and time and $x,y,t$ are the space and time locations where $g()$ applied. \textbf{Pool2} is a global sum pooling: 
\[
g(x) = \sum_{x, y, t} x_{x,y,t}.
\]

Using above key operators, the \textbf{IDT-HOG Net} represents the procedure of generating a KMeans-based bag of words (BoW) encoded HOG feature from stacked frames. At the first stage, the stacked frames are convolved with two gradient filters followed by 8 oriented binning filters and one spatio-temporal sum pooling. During the second stage, the descriptors from the first stage are convolved with K binning filters learned using KMeans and pooled together afterwards. The \textbf{IDT-HOF Net} and \textbf{IDT-MBH Net} represent the procedures of generating  KMeans encoded HOF and MBH features, respectively, from stacked optical flows. \textbf{IDT-MBH Net} is similar to the \textbf{IDT-HOG Net} except taking optical flows as inputs instead of pixels. \textbf{IDT-HOF Net} removes \textbf{Conv1} and using 9 oriented binning filters instead of 8.  Note that although we use KMeans encoding as an example, other encoding methods such as Fisher Vector and VLAD have similar procedures (\cite{sydorov2014deep,richardbow}). For simplicity, we leave out the feature detection step, which can be viewed as another convolution with binary activation function.  The main strength of this pipeline is that it is computationally efficient because of the layer-wise training and does not need labels to train the feature extraction module due to the objection of reconstructing the data itself. Its limitations lie in the first stage of the structure (dashed \textcolor{red}{red} box) in which it uses fixed parameters and structures for different sources of data.

\subsection{Comparison with CNN-based video features}

Needless to say CNNs employ convolution-pooling architecture.  In CNNs, the
non-linear activation is generally given by  $f(x) = \tanh(x)$, $f(x) =  (1 + e^{-x})^{-1}$ or $f(x) = \max(x, 0)$.  The pooling functions are local average or maximum pooling, for example,

\[
g_{x,y,t}(x) = \max_{j,l,m \in [1,d]} x_{xj,yl,tm },
\]
where $d$ is the pool size in space and time.
The parameters of the model are the filters $w$.  These are learned by
minimizing a loss function, typically defined by
\[
\min_w \sum_{i=1}^n||y_i - h(w, x_i)||^2 
\]
where $h(w,\cdot)$ is the full convolution-pooling architecture that takes $x$ as inputs. As can be noted above, the loss function requires the labels $y$ of the training data.

Comparing the above two procedures, it is clear that their differences are not so much structural, but rather in how to get the network parameters. With this understanding, we try to answer the question of how to design a video feature that balances efficiency and effectiveness. At first, one might try performing end-to-end training on the network structure in Figure \ref{fig:idt_net}. However, this training again requires labels and large computational resources. In addition, results from \cite{sydorov2014deep} and \cite{richardbow} show that directly applying end-to-end learning on the traditional handcrafted pipelines would not bring large performance gains. So instead we keep the stage-wise unsupervised training to avoid the costly labeling and training. We address the limitations of handcrafted features by proposing two-stream ConvISA to replace the handcrafted filters and designing different network structures for pixel and optical flow modeling. 

\begin{figure}
\centering
\includegraphics[width=0.70\textwidth]{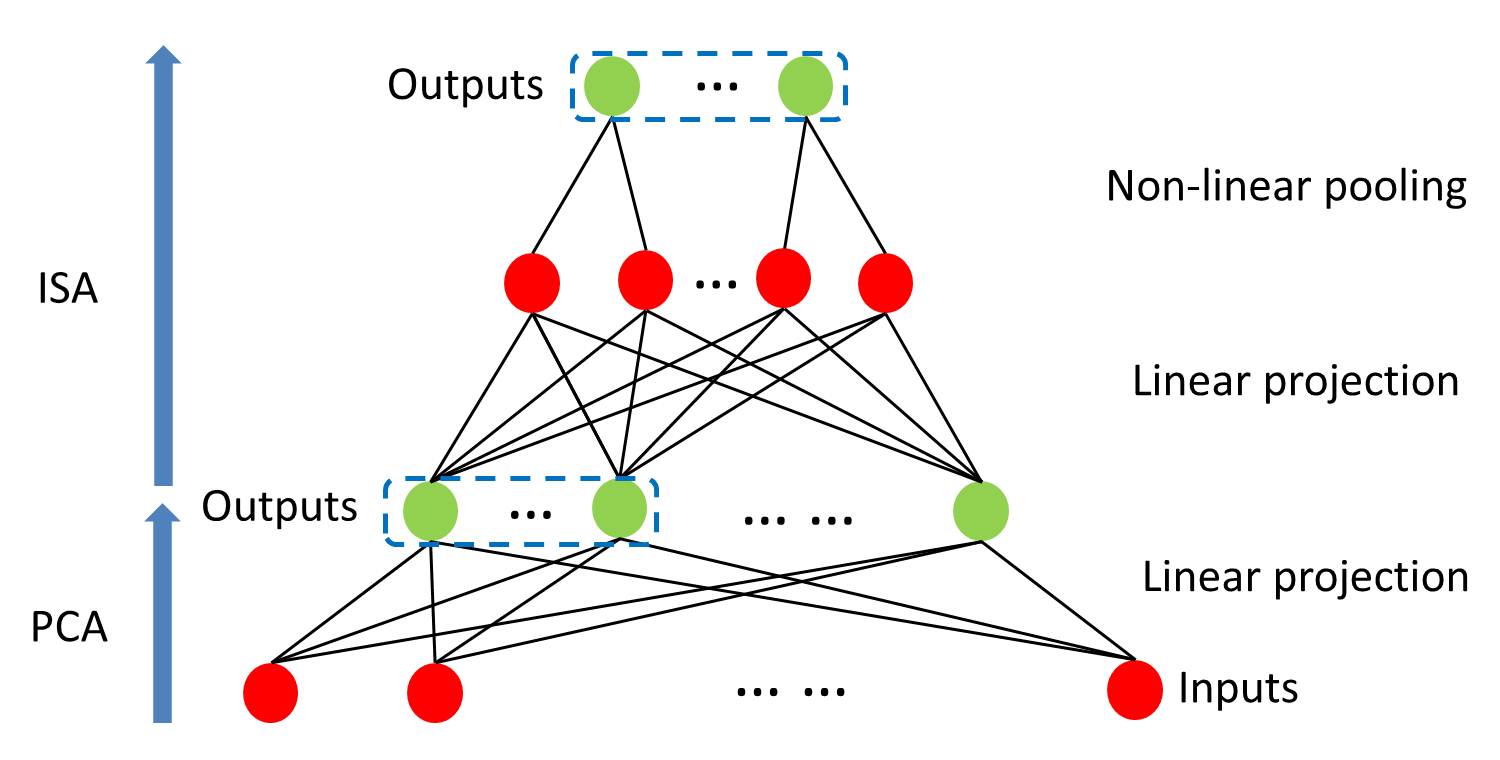}
\caption{ The neural network architecture of an ISA network with PCA preprocessing. The dashed \textcolor{blue}{blue} boxes represent the outputs of our model.}
\label{fig:isa_net}
\end{figure}

\section{Two-stream ConvISA}

In this section, we will describe two-stream ConvISA in detail and its structures for both appearance (pixel) and motion (optical flow) stream learning (\cite{simonyan2014two}). 

As illustrated in Figure \ref{fig:isa_net}, an ISA  (\cite{hyvarinen2009natural}) is a unsupervised feature learning method that can be described as a two-layered network within convolution-pooling architecture with: $f(x) = x^2$ and  $g(x) = \sqrt{x}$. A more formal and precise definition of ISA and its connection to group sparsity optimization can be found in Appendix. 

Figure \ref{fig:rn} visualizes, in both original and frequency domains, some example filters learned from ISA and PCA models. 
As illustrated in Figure \ref{fig:fb} and \ref{fig:fd} , the ISA model learns more complex filters that capture higher-frequency information. Furthermore, the pooling operation makes ISA lose information. To compensate for this information loss and capture more low-frequency information, we combine the outputs of ISA with an equivalent number of top outputs from PCA. Our enhanced method, denoted by ISA+, significantly outperforms individual PCA or ISA model. Since the PCA preprocessing is a default requirement for ISA, this improvement almost free.

\begin{figure}
\centering
    \begin{subfigure}[b]{0.4\textwidth}
                \includegraphics[width=\textwidth]{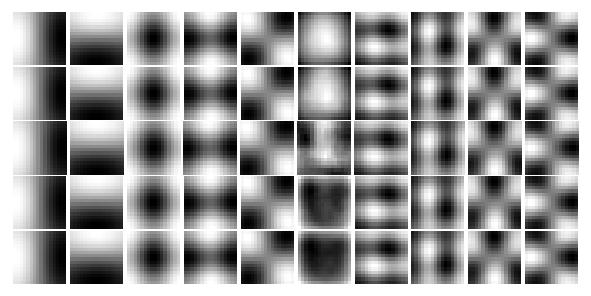}
                    \caption{PCA filters}
                    \label{fig:fa}
    \end{subfigure}
    \begin{subfigure}[b]{0.4\textwidth}
                \includegraphics[width=\textwidth]{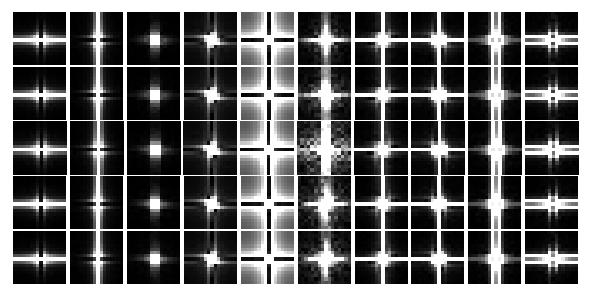}
                    \caption{PCA filters in frequency domain}
                    \label{fig:fb}
    \end{subfigure}
     \begin{subfigure}[b]{0.4\textwidth}
                \includegraphics[width=\textwidth]{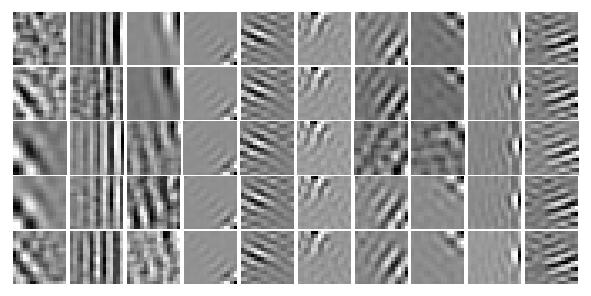}
                    \caption{ISA filters}
                    \label{fig:fc}
    \end{subfigure}
        \begin{subfigure}[b]{0.4\textwidth}
                \includegraphics[width=\textwidth]{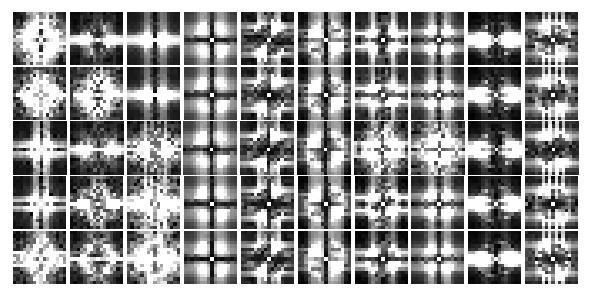}
                    \caption{ISA filters in frequency domain}
                    \label{fig:fd}
    \end{subfigure}
\caption{Example filters learned from our ConvISA model. For all figures, y-axis represents same component at different time step and x-axis represents different components. In frequency visualization, zero-frequency component are centered in the figure.}
\label{fig:rn}
\end{figure}

\begin{figure}
\centering
   \begin{subfigure}[b]{0.4\textwidth}
                \includegraphics[width=\textwidth]{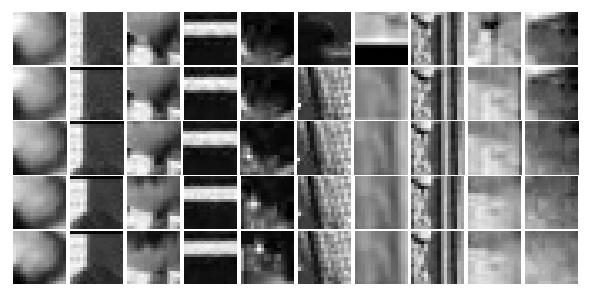}
                    \caption{Example image patches}
                    \label{fig:pa}
    \end{subfigure}
    \begin{subfigure}[b]{0.4\textwidth}
                \includegraphics[width=\textwidth]{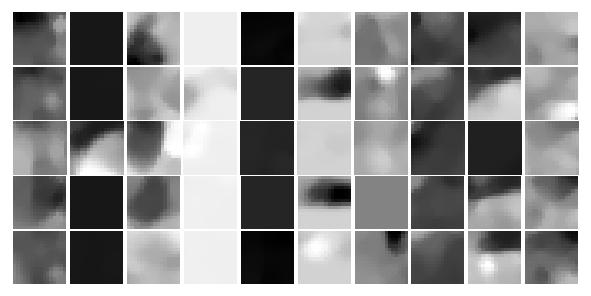}
                    \caption{Example optical flow patches}
                    \label{fig:pb}
    \end{subfigure}
    \begin{subfigure}[b]{0.4\textwidth}
                \includegraphics[width=\textwidth]{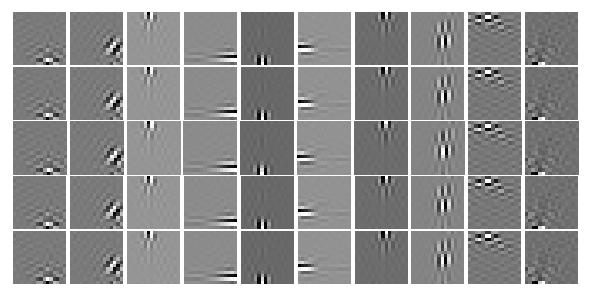}
                    \caption{Image filters learned from single frames}
                    \label{fig:pc}
    \end{subfigure}
    \begin{subfigure}[b]{0.4\textwidth}
                \includegraphics[width=\textwidth]{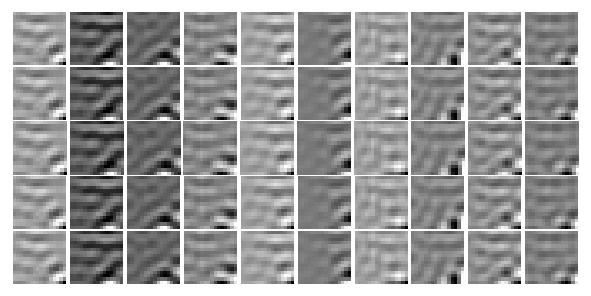}
                    \caption{Flow filters learned from single frames}
                    \label{fig:pd}
    \end{subfigure}
     \begin{subfigure}[b]{0.4\textwidth}
                \includegraphics[width=\textwidth]{img/grey_isa_tp5}
                    \caption{Image filters learned from 5 frame stacks}
                    \label{fig:pg}
    \end{subfigure}
        \begin{subfigure}[b]{0.4\textwidth}
                \includegraphics[width=\textwidth]{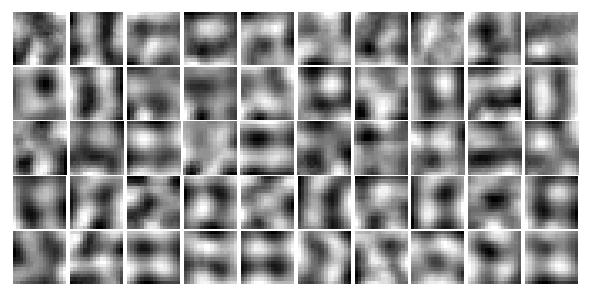}
                    \caption{Flow filters learned from 5 frame stackes}
                    \label{fig:ph}
    \end{subfigure}
\caption{Example inputs and filters learned from our ISA models. For each figure, the y-axis represents same component at different time steps and  the x-axis represents different component expect on Figure \ref{fig:pc} and \ref{fig:pd}, we replicate the filters 5 times for visualization purpose. }
\label{fig:tp-tp}
\end{figure}

To address the problem of handcrafted features having the same network structure for different data sources, we design different network structures for pixel and optical flow data. Our learned descriptor for appearance stream, dubbed LOP, is learned by directly applying a ConvISA model to a stack of video frames and implicitly learning temporal pooling. Our learned descriptor of motion stream, denoted as LOF, is trained by applying a ConvISA model to each individual optical flow frame and explicitly performing a temporal pooling afterwards. This difference of the network structure is because of our observation that pixel data has high temporal correlation while optical flow data often has much less temporal correlation due to the estimation error. As a result, it is much easier to learn temporally consistent appearance filters than temporally consistent motion filters. As shown in Figure \ref{fig:tp-tp}, in which we show some example images and optical flow patches and the filters learned in both structures. From Figure \ref{fig:pa} and \ref{fig:pb}, it is clear that image patches are consistent across frames while optical flow patches have large temporal variation.  Quantitatively, we estimate a 0.8014 temporal correlation for pixels sampled from HMDB51 tracklets while only 0.2808 for the same measurement on optical flow. As a result, the learned appearance filters (Figure \ref{fig:pg}) are temporally consistent and the learned flow filters are quite chaotic (Figure \ref{fig:ph}). Since temporally similar filters have the effect of pooling, these results show that temporal pooling is necessary for learning video features. To discriminate this implicit temporal pooling from the explicit temporal pooling for optical flow data, we call it temporal projection.

\section{Experiments}
\label{sec:exp}

In the following section, we first compare our learned descriptors to the state-of-the-art video descriptors. We then show that our ISA+ model performs significantly better than either one of the single models. After that, we demonstrate that custom designed network structures for pixel and optical flow data are necessary. We conduct experiments on benchmark action recognition datasets of HMDB51 and UCF101 datasets. 

\subsection{Datasets}

The HMDB51 dataset (\cite{kuehne2011hmdb}) has 51 action classes and 6766 video clips extracted from digitized movies and YouTube. \cite{kuehne2011hmdb} provides both original videos and stabilized ones. We only use original videos in this paper. As in \cite{kuehne2011hmdb}, Mean accuracy (MAcc) is used for evaluation. 

The UCF101 dataset (\cite{soomro2012ucf101}) has 101 action classes spanning over 13320 YouTube videos clips. We use the standard splits with training and testing videos provided by \cite{soomro2012ucf101}. We report MAcc as in the original papers.

\subsection{Experimental settings}
As in \cite{wang2013action}, IDT features are extracted using 15 frame tracking, camera motion stabilization and RootSIFT normalization and described by Trajectory, HOG, HOF, MBH, LOP and LOF descriptors. Two-stream ConvISA models are trained on 200000 IDT trackets for each stream of data. For both PCA and ISA, we keep the filter size the same as in the handcrafted descriptors and use a pooling size of 10 for ISA.  Another PCA is used to reduce the dimensionality of the resulting descriptors by a factor of two. For Fisher Vector encoding,  we map the raw descriptors into a  Gaussian Mixture Model with 256 Gaussians trained from a set of 256000 randomly sampled  data points. After encoding, we attach the normalized space-time locaiton information to the encoded descriptors as suggested in \cite{lan2014beyond}. Power and $\ell_{2}$ normalization are also used before concatenating different types of descriptors into a video based representation. For classification, we use a linear SVM classifier with a fixed $C=100$ as recommended by \cite{wang2013action} and the one-versus-all approach is used for multi-class classification scenario.  

\subsection{Performance comparison of individual descriptors}

In Table \ref{tab:descriptors}, we compare our learned descriptors LOG and LOF to the state-of-the-art video descriptors including handcrafted descriptors from IDT and spatial and temporal CNNs from \cite{simonyan2014two}. On the appearance descriptors, an impressive performance improvement of more than 10\% over HOG, from $42.0\%$ to $52.4\%$ is achieved by LOG on HMDB51. For UCF101, LOG also gets more than $7\%$ improvement over HOG and Spatial CNNs despite the fact that Spatial CNNs utilize additional training data. The same trend can be observed on motion descriptors. LOF outperforms other state-of-the-art descriptors by more than $4\%$ on HMDB51 and more than $3\%$ on UCF101. These results demonstrate that, with the same structure, the performance of handcrafted pipeline can be significantly boosted by learned descriptors.

\begin{table}[!h]
\centering
\begin{tabular}{|c |c c c | c c  c  c|}
\hline
    & \multicolumn{3}{c|}{Appearance Descriptors} & \multicolumn{4}{c|}{Motion Descriptors} \\
\hline
                & HOG & Spatial CNNs\footnote{The first split results from \cite{simonyan2014two}, pretrained on Imagenet and trained HMDB51 and UCF101 together (multi-task learning)} & LOG  &  HOF & MBH & Temporal CNNs \footnote{The first split results from \cite{simonyan2014two}, trained HMDB51 and UCF101 together (multi-task learning)}& LOF \\
                 \hline
HMDB51           & 42.0\%& N.A. & \textbf{52.4\%} &  49.8\% & 52.4\%  &  55.4\% &  \textbf{59.5\%} \\
\hline
UCF101           & 72.4\% & 72.8\% & \textbf{80.0\%} & 74.6\% & 81.4\% & 81.2\%  & \textbf{84.8\%}\\
  \hline
\end{tabular}
\caption{Comparison of our proposed descriptors to the state-of-the-art descriptors. }
\label{tab:descriptors}
\end{table}

\subsection{ISA+ is better than individual PCA or ISA models}

Table \ref{tab:ConvISA} compares our ISA+ model with individual PCA and ISA models. First, comparing PCA and ISA, we observe that, surprisingly, a simple PCA model can get similar results to a much more complex ISA model. These results demonstrate that PCA can learn good features when we require the number of features to be small. By combining the PCA and ISA outputs, we get more than $3\%$ improvement on HMDB51 and $2\%$ on UCF101. It should be noted that these improvements are on the combined results of appearance and motion models where the potential for improvement is less.

\begin{table}[!h]
\centering
\begin{tabular}{|c c c c|}
\hline
                & PCA & ISA & ISA+\\
                 \hline
HMDB51           & 58.1\% &  58.4\% & \textbf{61.5\%}\\
\hline
UCF101           & 85.9\% &  86.2\% & \textbf{88.3\%} \\
  \hline
\end{tabular}
\caption{Comparison of different unsupervised feature learning methods.}
\label{tab:ConvISA}
\end{table}

\subsection{Temporal projection versus temporal pooling}

In Table \ref{tab:tp-tp}, we compare the results of temporal projection and temporal pooling. As evidenced by the results of both datasets, for appearance modeling, it is better to do temporal projection than temporal pooling, and for motion modeling, temporal pooling performs much better than temporal projection. Furthermore, if we compare the single frame image filters (Figure \ref{fig:pc}) to the filters learned using 5 frame stacks (Figure \ref{fig:pg}) , we can see that adding temporal variation can help to learn more complex filters.  A potential improvement, therefore, is to explicitly enforce temporal coherence for optical flow filter learning and learn the temporal pooling for optical flow data.

\begin{table}[!h]
\centering
\begin{tabular}{|c| c c| c c|}
\hline
                & \multicolumn{2}{c|}{LOG} & \multicolumn{2}{c|}{ LOF}\\
                \hline
                & Temporal Projection & Temporal Pooling & Temporal Projection & Temporal Pooling\\
                 \hline
HMDB51          &  \textbf{52.4\%} & 44.3\%   & 46.2\% & \textbf{59.5\%}\\
\hline
UCF101           &  \textbf{80.0\%} & 73.5\%   & 79.6\% & \textbf{84.8\%} \\
  \hline
\end{tabular}
\caption{Comparison of temporal projection and temporal pooling.}
\label{tab:tp-tp}
\end{table}

\section{Conclusions}

Contrary to the current trend of learning video features using end-to-end deep CNNs, which is computationally intensive and label demanding, we propose in this paper to revisit the traditional local feature pipeline and combine the merits of both hancrafted and CNN approaches. As an example, we present a video feature learning algorithm that has better performance, lower computational expensive than current state-of-the-art methods and does not require labels. We show that filters learned in an unsupervised fashion , when incorporates in a convolution-pooling structure, can outperform supervised end-to-end networks. This result serves as a reminder that the design choices in handcrafted features may still have many useful properties which could be potentially incorporated into future deep action recognition networks. Future work would be explicitly enforcing temporal consistency for optical flow feature learning and developing a deeper and better unsupervised learning method. We would also like to explore end-to-end fine-tuning given the unsupervised learned networks, which is less expensive than training from scratch.

\bibliography{iclr2016_conference}
\bibliographystyle{iclr2016_conference}

\section{Appendix}

ISA can be described as a two-layered network. Specifically, let matrix $W\in\mathbb{R}^{m\times n}$ and matrix $V\in\mathbb{R}^{d\times m}$ denote the parameters of the first and second layers of ISA respectively. $n$ is the dimension of the inputs and $d$ is the dimension of outputs. $m$ is the number of latent variables between the first layer and the second layer. Typically $d\leq m\leq n$. The matrix $W$ is learned from data with orthogonal constraint $WW{}^{\top}=I$. Therefore we call $W$ the projection matrix. The matrix $V$ is given by the network structure to group the output variables of the first
layer. $V_{ij}=1$ if the $j$-th output variable of the first layer is in the $i$-th group, otherwise $V_{ij}=0$. Therefore we call $V$ the grouping matrix. Given an input pattern $X^{t}\in\mathbb{R}^{n}$, the activation of $i$-th output unit of the second layer is $p_{i}(X^{t};W,V)$
defined by 
\begin{align}
p_{i}(X^{t};W,V)\triangleq\sqrt{\sum_{k=1}^{m}V_{ik}(\sum_{j=1}^{n}W_{kj}X_{j}^{t})^{2}}\ .
\end{align}
 ISA enforces the activation of the output unit to be sparse.
To achieve the sparse activation, it minimizes the following loss function defined
on $T$ training instances:
\begin{align}
\min_{W}\quad & \sum_{t=1}^{T}\sum_{i=1}^{d}p_{i}(X^{t};W,V,)\label{eq:ISA-network-formu}\\
\mathrm{s.t.}\quad & WW{}^{\top}=I\ .\nonumber 
\end{align}

 Another way to interpret ISA is from sparse coding framework. Let
$\mathcal{G}=[\mathcal{G}_{1},\mathcal{G}_{2},\cdots,\mathcal{G}_{d}]$
denote the variable group indexes defined by $V$, that is, $j\in\mathcal{G}_{i}$
if and only if $V_{i,j}=1$. $|\mathcal{G}_i|$ defines group size, which is generally the same across groups. 

As in group LASSO (\cite{friedman2010note}), for
any vector $\boldsymbol{a}\in\mathbb{R}^{m}$, we defined the group
$\ell_{1}$-norm $\|\boldsymbol{a}\|_{\mathcal{G},1}$ as
\[
\|\boldsymbol{a}\|_{\mathcal{G},1}\triangleq\sum_{i=1}^{d}\sqrt{\sum_{j\in\mathcal{G}_{i}}\boldsymbol{a}_{j}^{2}}\ .
\]
 We can write $p_{i}(X^{t};W,V)$ as
\[
p_{i}(X^{t};W,V)=\|WX^{t}\|_{\mathcal{G},1}\ .
\]
 Denote $\boldsymbol{\alpha}_{t}=WX^{t}$, since $WW{}^{\top}=I$,
we have 
\[
X^{t}=W^{\dagger}\boldsymbol{\alpha}_{t}\ ,
\]
 where $W^{\dagger}$ is the Moore\textendash Penrose pseudo inverse
of $W$. Eq. (\ref{eq:ISA-network-formu}) can be re-formulated as
a sparse coding method that 
\begin{align}
\min_{W, \boldsymbol{\alpha_t}}\quad & \sum_{t=1}^{T}\|\boldsymbol{\alpha}_{t}\|_{\mathcal{G},1}\label{eq:ISA-sparse-coding-formu}\\
\mathrm{s.t.}\quad & (W^{\dagger}){}^{\top}W{}^{\dagger}=I\nonumber 
 & X^{t}=W^{\dagger}\boldsymbol{\alpha}_{t}\nonumber 
\end{align}
 Based on Eq. (\ref{eq:ISA-sparse-coding-formu}), ISA is essentially searching
a group-sparse representation $\boldsymbol{\alpha}_{t}$ of the input
signal $X_{t}$. The matrix $W^{\dagger}$ is the dictionaries of
sparse coding. The orthogonal constraint of $W^{\dagger}$ makes the
learned components maximally independent.

\end{document}